\pdfoutput=1

\documentclass[11pt]{article}

\usepackage{naacl2021}

\usepackage{times}
\usepackage{latexsym}

\usepackage[T1]{fontenc}

\usepackage[utf8]{inputenc}

\usepackage{microtype}

\usepackage{amsmath,amsfonts,bm}

\def\eqref#1{equation~\ref{#1}}

\def\1{\bm{1}}

\def\rvw{{\mathbf{w}}}
\def\rvx{{\mathbf{x}}}

\def\rvz{{\mathbf{z}}}

\DeclareMathAlphabet{\mathsfit}{\encodingdefault}{\sfdefault}{m}{sl}
\SetMathAlphabet{\mathsfit}{bold}{\encodingdefault}{\sfdefault}{bx}{n}

\newcommand{\ie}{\textit{i.e.}}
\newcommand{\eg}{\textit{e.g.}}
\usepackage{graphicx}
\usepackage{booktabs}
\usepackage{wrapfig}
\usepackage{subfig}
\usepackage{colortbl}
\usepackage{comment}
\usepackage{footnote}
\usepackage[outline]{contour}
\usepackage{multirow}

\definecolor{col1}{RGB}{68,133,62}
\definecolor{col3}{RGB}{68,133,62}
\definecolor{white}{RGB}{255,255,255}
\definecolor{na}{gray}{0.9}

\newcommand{\greenarrowup}{{\color{col1}\contourlength{0.05em}\contour{col1}{$\nearrow $}}}

\newcommand{\greenarrowdown}{{\color{col3}\contourlength{0.05em}\contour{col3}{$\searrow $}}}

\newcommand{\hiddenarrow}{{\color{white}\contourlength{0.05em}\contour{white}{$\rightarrow $}}}

\newcommand{\method}[0]{UBM}
\newcommand{\methodlong}[0]{Upstream Bias Mitigation for Downstream Fine-Tuning}

\newcommand{\methodlongwacr}[0]{Upstream Bias Mitigation (UBM) for Downstream Fine-Tuning}

\title{On Transferability of Bias Mitigation Effects in Language\\ Model Fine-Tuning}

\author{Xisen Jin\raisebox{3pt}{$\S$}, Francesco Barbieri\raisebox{3pt}{$\dagger$},  Brendan Kennedy\raisebox{3pt}{$\S$}, Aida Mostafazadeh Davani\raisebox{3pt}{$\S$}, \\
\textbf{Leonardo Neves\raisebox{3pt}{$\dagger$}, Xiang Ren\raisebox{3pt}{$\S$}} \\
\raisebox{1pt}{$\S$}University of Southern California\\
\raisebox{1pt}{$\dagger$}Snap Inc.\\ 
\texttt{\{xisenjin, mostafaz, btkenned, xiangren\}@usc.edu} \\
\texttt{\{fbarbieri, lneves\}@snap.com} \\
}

\begin{document}

\maketitle

\begin{abstract}

Fine-tuned language models have been shown to exhibit biases against protected groups in a host of modeling tasks such as text classification and coreference resolution. 
Previous works focus on detecting these biases, reducing bias in data representations, and using auxiliary training objectives to mitigate bias during fine-tuning.
Although these techniques achieve bias reduction for the task and domain at hand, the effects of bias mitigation may not directly transfer to new tasks, requiring additional data collection and customized annotation of sensitive attributes, and re-evaluation of appropriate fairness metrics.
We explore the feasibility and benefits of \textit{upstream bias mitigation} (\method) for reducing bias on downstream tasks, by first applying bias mitigation to an upstream model through fine-tuning and subsequently using it for downstream fine-tuning. 
We find, in extensive experiments across hate speech detection, toxicity detection, occupation prediction, and coreference resolution tasks over various bias factors, that the effects of~\method~are indeed transferable to new downstream tasks or domains via fine-tuning, creating less biased downstream models than directly fine-tuning on the downstream task or transferring from a vanilla upstream model.
Though challenges remain, we show that~\method~promises more efficient and accessible bias mitigation in LM fine-tuning.\footnote{Code and data: \url{https://github.com/INK-USC/Upstream-Bias-Mitigation}}\footnote{The work was partially done when Xisen Jin was an intern at Snap Inc.} 


\end{abstract}

\section{Introduction}

The practice of fine-tuning pretrained language models (PTLMs or LMs), such as BERT  \citep{Devlin2019BERTPO}, has improved prediction performance in a wide range of NLP tasks. However, fine-tuned LMs may exhibit biases against certain protected groups (e.g., gender and ethnic minorities), as models may learn to associate certain features with positive or negative labels spuriously~\citep{Dixon2018MeasuringAM}, or propagate bias encoded in PTLMs to downstream classifiers~\citep{Caliskan2017SemanticsDA, Bolukbasi2016ManIT}.
Among many examples,~\citet{Kurita2019MeasuringBI} demonstrates gender-bias in the pronoun resolution task when models are trained using BERT embeddings, and~\citet{Kennedy2020ContextualizingHS} shows that hate speech classifiers fine-tuned from BERT result in more frequent false positive predictions for certain group identifier mentions (\eg, ``\textit{muslim}'', ``\textit{black}'').

\begin{figure}
\vspace{-0.3cm}
    \centering
    \includegraphics[width=\linewidth]{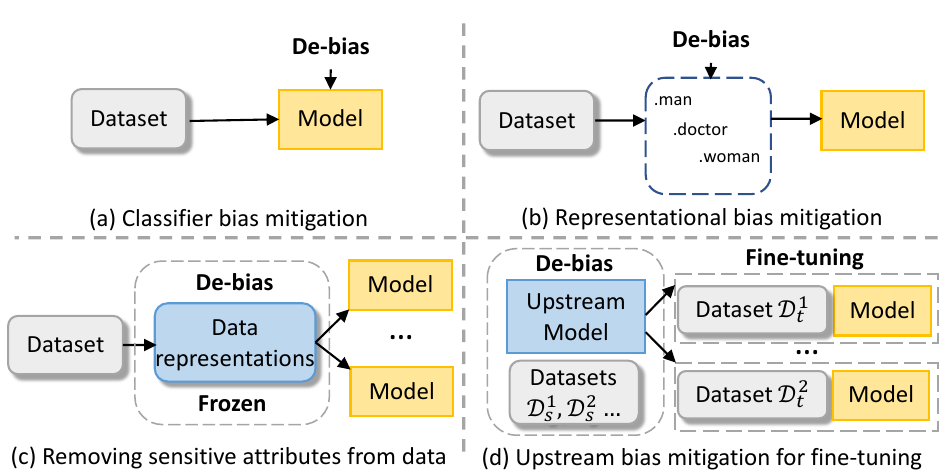}
    \caption{\textbf{Comparison between the focus of our study (d) and previous works (a,b,c).} We study the viability of obtaining an upstream model that could reduce bias in a number of downstream classifiers when fine-tuned.}
    \label{fig:intro_related}
\vspace{-0.4cm}
\end{figure}

Approaches for bias mitigation are mostly applied during fine-tuning to reduce bias in a specific downstream task or dataset~\citep{Park2018ReducingGB,Zhang2018MitigatingUB, Beutel2017DataDA} (see Fig.~\ref{fig:intro_related} (a)). 
For example, data augmentation approaches reduce the influence of spurious features in the original dataset~\citep{Dixon2018MeasuringAM, Zhao2018Gender, Park2018ReducingGB}, and adversarial learning approaches generate debiased data representations that are exclusive to the downstream model \citep{Kumar2019TopicsTA, Zhang2018MitigatingUB}.
These techniques act on biases particular to the given dataset, domain, or task, and require new bias mitigation when switching to a new downstream task or dataset. This can require auxiliary training objectives, the definition of task-specific fairness metrics, the annotation of bias attributes (\eg, identifying African American Vernacular English), and the collection of users' demographic data.
These drawbacks make bias mitigation inaccessible to the growing community, fine-tuning LMs to new datasets and tasks.



In contrast, we investigate initially mitigating bias while fine-tuning an ``upstream'' model in one or more upstream datasets, and subsequently achieving reduced bias when fine-tuning for downstream applications (Fig.~\ref{fig:intro_related} (d)), so that bias mitigation is no longer required in downstream training.
Similar to transfer learning for enhancing predictive performance in common setups~\cite{Pan2010ASO, Dai2015SemisupervisedSL}, we suggest that 
LMs that undergo bias mitigation acquire inductive bias that is helpful for reducing harmful biases when fine-tuned on new domains and tasks.
In four tasks with known bias factors --- hate speech detection, toxicity detection, occupation prediction from short bios, and coreference resolution --- we explore whether upstream bias mitigation of a LM followed by downstream fine-tuning reduces bias for the downstream model.
Though previous work has addressed biases in frozen PTLM or word embeddings~\cite{Bolukbasi2016ManIT,Zhou2019ExaminingGB,Bhardwaj2020InvestigatingGB, Liang2020TowardsDS, Ravfogel2020NullIO}, for example by measuring associations between gender and occupations in an embedding space, they do not study their effect on downstream classifiers (Fig.~\ref{fig:intro_related} (b)), while some of them study the effects while keeping the embeddings frozen~\cite{Zhao2019GenderBI, Kurita2019MeasuringBI, Prost2019DebiasingEF}. Bias in these frozen representations can also be directly corrected by removing associations between feature and sensitive attributes~\cite{Elazar2018AdversarialRO, Madras2018LearningAF} (Fig.~\ref{fig:intro_related} (c)), but this does not allow predictions to be generated for new data. 


Our experiments address the following research questions: (a) whether mitigating a single bias factor in the upstream stage is maintained when fine-tuning on new examples from the same domain and task, (b) whether transfer is viable when the downstream domains and tasks are different from the upstream model,
and (c) whether we can address multiple kinds of bias with a single upstream model.
We perform these experiments under a generic transfer learning framework, noted as \methodlongwacr~for convenience, which consists of two stages: first, in the \textit{upstream bias mitigation stage}, a LM is fine-tuned with bias mitigation objectives on one or several ``upstream'' tasks, and subsequently the classification layer is re-initialized; then, in the \textit{downstream fine-tuning stage} the encoder from the upstream model, jointly with the new classification layer, are again fine-tuned on a downstream task without additional bias mitigation steps. Using six datasets with previously recognized bias factors, our analysis show overall positive results for the questions above; still, there are challenges remaining to stabilize the results of bias mitigation in challenging setups, \eg, the multi-bias factor setting.

Our contributions are summarized as follows: (1) we propose a new research direction for mitigating bias in fine-tuned models; (2) we perform extensive experiments to study the viability of the upstream bias mitigation framework in various settings; (3) we demonstrate the effectiveness of this research direction, motivating further improvements, tests, and applications.

\section{Exploring the Transferability of Bias Mitigation Effects}
\vspace{-0.1cm}
We consider biases against protected groups in classifiers fined-tuned from LMs.
In our present analysis, bias is defined as disparate model performance on different subsets of data which are associated with different demographic groups (\eg, instances that mention or are generated by different social groups)~\citep{Blodgett2020LanguageI}. 
Our evaluation of bias aligns with the definition of equalized odds and equal opportunities~\cite{Hardt2016EqualityOO} in previous works of fairness in machine learning. 

Here, we first outline our experimental setup for exploring the transferability of bias mitigation effects, in which we detail the process of applying~\method~and pose three key research questions (section~\ref{ssec:rq}). 
We follow by introducing the bias factors studied and the corresponding classification tasks and datasets (section~\ref{ssec:bf_d}), and our evaluation protocols and metrics (section~\ref{ssec:ep_m}).

\subsection{Experiment Setups of~\method}
\label{ssec:rq}

\begin{figure*}[t!]
    \vspace{-0.2cm}
    \centering
    \includegraphics[width=0.90\textwidth]{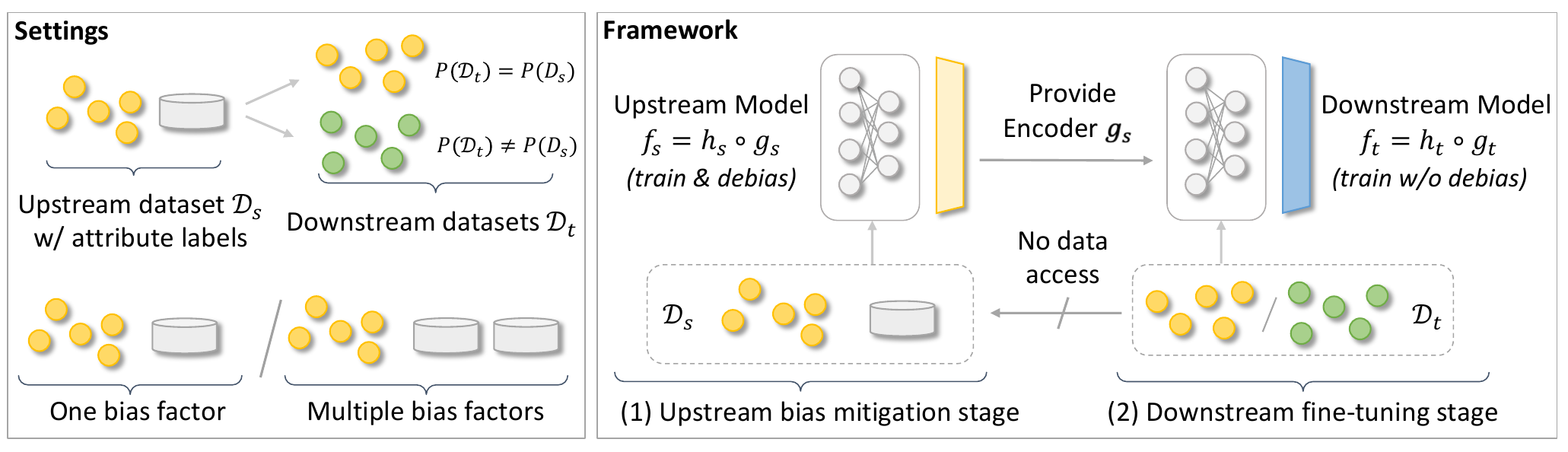}
    \vspace{-0.1cm}
    \caption{\textbf{Experiment setups to study \methodlongwacr}. We consider the settings with the same or different upstream and downstream domains and tasks, while addressing one or more bias factors (\eg, both dialect bias and gender bias). The framework consists of two stages: (1) an upstream (source) model $f_s=h_s \circ g_s$ is trained with bias mitigation algorithms and (2) the encoder $g_s$ is transferred to the downstream (target) model $f_t$ for fine-tuning.}
    \vspace{-0.2cm}
    \label{fig:framework}
\end{figure*}


Our goal is to evaluate the transferability of bias mitigation effects for \textit{one} or \textit{multiple} bias factors in downstream fine-tuned models. We follow an~\methodlongwacr~procedure, pictured in Figure~\ref{fig:framework}. 
First, in the \textit{Upstream Bias Mitigation} phase, an upstream (source) model $f_s=h_s \circ g_s$, composed of a text encoder $g_s$ and a classifier head $h_s$, is trained on one or more upstream datasets $\mathcal{D}_s$ with bias mitigation algorithms. The encoder $g_s$ is to be transferred to downstream (target) domains and tasks while the classifier head $h_s$ is discarded. Then, in the \textit{Downstream Fine-Tuning} phase, the downstream model $f_t = h_t \circ g_t$ utilizes $g_s$ to initialize the encoder weights and is fine-tuned for prediction performance without bias mitigation approaches on downstream datasets $\mathcal{D}_t$. 

This \method~process is applied in three settings, summarized below, which each contribute to evaluating the transferability of bias mitigation effects. 

\smallskip
\noindent \textbf{1. Fine-Tuning on the Same Distribution.} In the simplest setting, we fine-tune the downstream model over new examples from the same data distribution as the upstream model. In practice, each dataset is split into two halves, with one used for upstream bias mitigation and the other for downstream fine-tuning. 

\smallskip
\noindent \textbf{2. Cross-Domain and Cross-Task Fine-Tuning.} 
Similar to how LMs are fine-tuned for various tasks and domains, in a more practical setup, we test whether transfer of bias mitigation effects is viable across domains and tasks. 
To achieve this, we apply bias mitigation while fine-tuning a LM on one dataset and perform fine-tuning on another.

\smallskip
\noindent \textbf{3. Multiple Bias Factors.} 
In the most challenging setup, we train a single upstream model to address multiple bias factors (\eg, both dialect bias and gender bias). Such upstream models can be trained with multi-task learning (\ie, jointly training over multiple datasets with shared encoder $g$ but different classifier heads $h$) while mitigating multiple kinds of bias. Subsequently, the resulting upstream model is transferred to downstream models as before. This is a key test of \method's viability for widespread application.

\vspace{-0.1cm}

\subsection{Bias Factors and Datasets}
\label{ssec:bf_d}

To ensure our analysis holds true for a variety of domains, tasks, and bias factors, we experiment with three different bias factors studied in previous research along with six different datasets (also summarized in Table~\ref{tab:dataset_summary}), described below.  

\smallskip
\noindent \textbf{Group Identifier Bias.} This bias refers to higher false positive rates of hate speech predictions for sentences containing specific group identifiers, which is harmful to protected groups by misclassifying innocuous text (\eg, ``I am a Muslim'') as hate speech. We include two datasets for study, namely the Gab Hate Corpus~\citep[GHC;][]{kennedy2018gab} and the Stormfront corpus~\citep{de2018hate}. Both datasets contain binary labels for hate and non-hate instances, though with differences in the labeling schemas and domains.

\begin{table}[t!]
\centering
\scalebox{0.58}{
\begin{tabular}{@{}lcc@{}}
\toprule
\textbf{Dataset}       & \textbf{Prediction Task}        & \textbf{Bias}         \\ \midrule
{GHC}~\cite{kennedy2018gab}         & Hate        & Group Identifier     \\
{Stormfront}~\cite{de2018hate}    & Hate        & Group Identifier     \\
{DWMW}~\cite{davidson2017automated}          & Toxicity    & AAVE Dialect \\
{FDCL}~\cite{Founta2018LargeSC}          & Toxicity    & AAVE Dialect \\
{BiasBios}~\cite{de2019bias}      & Occupation  & Gender Stereotyping       \\
OntoNotes 5.0~\cite{weischedel2013ontonotes} & Coreference & Gender Stereotyping     \\ \bottomrule
\end{tabular}
}
\vspace{-0.2cm}
\caption{Summary of tasks and bias included for study.}
\label{tab:dataset_summary}
\vspace{-0.4cm}
\end{table}

\smallskip
\noindent \textbf{AAVE Dialect Bias.}~\citet{Sap2019TheRO} show that offensive and hate speech classifiers yield a higher false positive rate on text written in African American Vernacular English (AAVE). This bias brings significant harm to the communities that uses AAVE, for example, by leading to the disproportionate removal of the text written in AAVE in social media platforms~\citep{Blodgett2020LanguageI}. 
We include two datasets for study: FDCL~\citep{Founta2018LargeSC} and DWMW~\citep{davidson2017automated}.
In both datasets, we treat \textit{abusive}, \textit{hateful} and \textit{spam} together as harmful outcomes (\ie, false positives for each are harmful) to compute false positive rates. Following~\citet{Sap2019TheRO}, we use an off-the-shelf AAVE dialect predictor~\citep{Blodgett2016DemographicDV} to identify examples written in AAVE.

\smallskip
\noindent \textbf{Gender Stereotypical Bias.}~
\citet{Zhao2018Gender} summarize a list of occupations that are prone to be stereotyped in practice, leading to coreference resolutions models and occupation prediction models having biases in performance in pro- and anti-stereotypical instances when trained on short bios. 
We train the coreference resolution model on the OntoNotes 5.0 dataset~\cite{weischedel2013ontonotes} and the occupation classifier on the BiasBios~\cite{de2019bias} dataset.

\subsection{Evaluation Protocol and Metrics}
\label{ssec:ep_m}

%

We evaluate the overall performance of the models on downstream tasks along with appropriate bias metrics for each bias factor, analyzed for each dataset and task in previous works.
We expect \method~to minimally affect classification performance while improving on bias metrics.

\smallskip
\noindent \textbf{Classification Performance.}
We report in-domain F1 scores for GHC, Stormfront, OntoNotes 5.0, and accuracy scores for FDCL, DWMW and BiasBios. Following~\citet{Zhang2018MitigatingUB}, for hate speech detection and toxicity detection datasets, we use the equal error rate (EER) threshold for prediction.

\smallskip
\noindent \textbf{Group Identifier Bias Metrics.} To evaluate group identifier bias, we evaluate false positive rate (FPR) differences, noted as FPRD, between examples mentioning one of 25 group identifiers provided by \citet{Kennedy2020ContextualizingHS} and the overall FPR. 
In addition, we followed \citet{Kennedy2020ContextualizingHS} in using a New York Times articles (NYT) corpus of 25$k$ non-hate sentences, each mentioning one of 25 group identifiers. This corpus specifically provides an opportunity to measure FPR---reported as ({NYT Acc.}), equivalent to $1-$FPR. Additionally, following the evaluation protocol of~\citet{Dixon2018MeasuringAM} and~\citet{Zhang2020DemographicsSN}, we incorporate the Identity Phrase Templates Test Sets (reported as IPTTS), which consists of $77k$ hate and non-hate examples mentioning group identifiers, generated with templates. Following these works, for IPTTS we compute FPRD as $\sum_{z} |\mathrm{FPR}_z - \mathrm{FPR}_{overall}|$, where $\textrm{FPR}_z$ is false positive rate on sentences with the group identifier $z$, and $\mathrm{FPR}_{overall}$ is the overall false positive rate. 

\smallskip
\noindent \textbf{AAVE Dialect Bias Metrics.} Given the sparsity of AAVE examples in the datasets and the noisy outputs of AAVE classifier~\citep{Blodgett2016DemographicDV}, we expect the in-domain FPRD metrics to be noisy. Therefore, following~\citet{Xia2020DemotingRB}, 
we incorporate the BROD~\citep{Blodgett2016DemographicDV} dataset, which is a large unlabeled collection of Twitter posts written in l. 
Since in practice only a small portion of texts are toxic or spam, we treat all examples from BROD as \textit{normal}, and report the accuracy (which equals $1-$FPR) on the dataset.

\smallskip
\noindent \textbf{Gender Stereotype Metrics.} We employ the WinoBias~\cite{Zhao2018Gender} dataset which provides opportunities to evaluate models on pro-stereotypical and anti-stereotypical coreference examples. We report the differences in F1 (F1-Diff) on two subsets of data. On occupation prediction, following~\citet{Ravfogel2020NullIO}, we report mean differences of true positive rate (TPR) differences in predicting each occupation for men and women.

\section{Method}





Here, we detail the particular bias mitigation algorithms used for implementing~\method, as well as the other baselines used for verifying the transferability of bias mitigation effects. 


\subsection{Implementations of~\method}
\label{ssec:bias_mitigation_approaches}

We implement~\method~with two different bias mitigation algorithms in the upstream bias mitigation phase: explanation regularization~\citep{Kennedy2020ContextualizingHS}, and adversarial de-biasing~\citep{Zhang2018MitigatingUB,Madras2018LearningAF,Xia2020DemotingRB}, denoted here as~\textbf{\method$_{reg}$} and~\textbf{\method$_{adv}$}, respectively.

\medskip
\noindent 
\textbf{\method~with Explanation Regularization.} Explanation regularization reduces importance placed on spurious surface patterns (\ie, words or phrases) during upstream model training. 
We apply \method$_{reg}$ to group identifier and AAVE dialect bias, where the set of spurious patterns are group identifiers and the most frequent words, from statistics of the dataset, used by AAVE speakers; we find explanation regularization not effective for gender bias. The importance of a surface pattern $w \in \mathcal{W}$ in the input $\rvx$, noted as $\phi(w, \rvx)$ is measured as the model prediction change when it is removed. The model is trained by optimizing the main learning objective $\ell$  while penalizing importance attributed to patterns $w \in \mathcal{W}$ that exist in the input $\rvx$. 
\begin{equation}
\label{eq:expl_reg}
    \min_{f} \quad \ell_{c} + \alpha \sum_{w \in \rvx \cap \mathcal{W} } ||\phi(w, \rvx)||^2,
\end{equation}
where $\alpha$ is a trade-off hyperparameter.

\medskip
\noindent 
\textbf{\method~ with Adversarial De-biasing.} In~\method$_{adv}$, the upstream model is trained with adversarial de-biasing techniques, so that sensitive attributes related to bias (\eg, the dialect of the sentence or the gender referenced in the sentence) cannot be predicted from the hidden representations $\rvz$ given by the encoder $g$. 
During training, an adversarial classifier head $h^{adv}$ is built upon the encoder and trained to predict sensitive attributes, while the encoder is optimized to prevent the adversarial classifier from success. Formally, the optimization objective is written as,
\begin{align}
    \min_{g,h} \max_{h_{adv}} \quad \ell_c + \ell_{adv}(h_{adv}  \circ g(\rvx), a),
\end{align}
where $a$ notes the ground truth sensitive attribute, and $\ell_{adv}$ is the cross entropy loss between the predicted sensitive attribute and the ground truth sensitive attribute. 

As mentioned in Sec.~\ref{ssec:rq}, upstream models can be trained to mitigate multiple bias factors with multi-task learning on multiple datasets. We separately apply bias mitigation algorithms for each dataset (sharing the same encoder) and note the algorithms applied in the subscript (\eg, \textbf{\method$_{reg+adv}$}).

\subsection{Other Baselines}
We compare~\method~with two families of methods.

\smallskip
\noindent \textbf{Methods without Bias Mitigation.} 
Two types of models were evaluated that did not address bias. First, the \textbf{Vanilla} model is a downstream classifier directly fine-tuned on downstream task from a LM (\eg, RoBERTa). Second, \textbf{Van-Transfer} is fine-tuned on upstream datasets without bias mitigation and fine-tuned on downstream datasets. 

\smallskip
\noindent \textbf{Downstream Bias Mitigation.}
For reference, we show the results of directly applying explanation regularization, noted as \textbf{Expl. Reg.}, or adversarial de-biasing, noted as \textbf{Adv. Learning}, during down-stream fine-tuning. In most cases, mitigating bias in downstream classifier should be the most effective way to reduce bias, though this is not always feasible in practice for reasons discussed above.

We also consider two simple baselines that could reduce bias in downstream models via heuristics. \textbf{Emb. Zero} zeros out the word embedding of spurious surface patterns (using the same word list as explanation regularization) in PTLMs before fine-tuning. We also include \textbf{Emb. Zero. Trans}, which zeros out embeddings of spurious surface patterns before fine-tuning from an upstream model. The method does not apply to cases where surface patterns related to bias (\eg, gendered pronouns) are crucial for prediction, \eg, coreference resolution.



\vspace{-0.0cm}
\section{Results}
\label{sec:exp}
\vspace{-0.1cm}

\begin{table}[t]
\vspace{-0.0cm}
\centering
\scalebox{0.60}{
\begin{tabular}{@{}l|cccc@{}}
\toprule
\textbf{} & \multicolumn{4}{c}{\textbf{GHC B}}                                                                                                                                                                                     \\ \midrule
\textbf{Metrics}           & \textbf{\begin{tabular}[c]{@{}c@{}}In-domain\\ F1 ($\uparrow$)\end{tabular}} & \textbf{\begin{tabular}[c]{@{}c@{}}In-domain\\ FPRD ($\downarrow$)\end{tabular}} & \textbf{\begin{tabular}[c]{@{}c@{}}IPTTS\\ FPRD ($\downarrow$)\end{tabular}} & \textbf{\begin{tabular}[c]{@{}c@{}}NYT\\ Acc ($\uparrow$)\end{tabular}}  \\ \midrule
                           & \multicolumn{4}{c}{\textit{Non-Transfer (GHC . B)}}                                                                                                                                                                                                                   \\ \midrule
\textbf{Vanilla}           & 37.91 $\pm$ 2.5                                                                & 35.64     $\pm$  2.2                                  &  21.50 $\pm$ 2.8                                                                                                          & 68.55 $\pm$ 20                                                        \\
\textbf{Expl. Reg.}        & \textbf{38.09 $\pm$ 2.7}                                                              & \textbf{18.68 $\pm$ 0.3}                                                                       & \textbf{4.82 $\pm$ 1.1}                                              & \textbf{84.05 $\pm$ 3.0}                                                                                    \\
 \midrule
\textbf{}                  & \multicolumn{4}{c}{\textit{GHC . A $\to$ GHC . B}}                                                                                                                                         \\ \midrule
\textbf{Van. Transfer}     & 42.41   $\pm$ 1.0                 & 37.44 $\pm$ 1.5                                         & 17.67    $\pm$ 2.1                                                                              & 75.35 $\pm$ 4.2                                              \\
\textbf{UBM$_{Reg}$}       & \textbf{43.79 $\pm$ 1.9}                                           & \textbf{34.34 $\pm$ 3.1}                                                       & \textbf{10.02 $\pm$ 1.1}                             & \textbf{81.40 $\pm$ 1.4}                                                        \\
 \bottomrule
\end{tabular}
}
\caption{\textbf{Same-domain and task UBM with a single bias factor}. The source datasets are noted before arrow $(\rightarrow)$. All metrics except In-domain F1 measure bias. See Table~\ref{tab:same_domain_full} in Appendix for complete results.}
\label{tab:same_domain_partial}
\vspace{-0.3cm}
\end{table}
\begin{table*}[t]
\vspace{-0.1cm}
\centering
\scalebox{0.50}{
\begin{tabular}{@{}l|cccc|cccc|cc|cc@{}}
\toprule
\textbf{Downstream dataset} & \multicolumn{4}{c|}{\textbf{GHC}}                                                                                                                                                                                                                                                                                        & \multicolumn{4}{c|}{\textbf{Stormfront}}                                                                                                                                                                                                                                                                                  & \multicolumn{2}{c|}{\textbf{DWMW}}                                                                                                                       & \multicolumn{2}{c}{\textbf{OntoNotes 5.0}}                                                                                                                        \\ \midrule
\textbf{Metrics}           & \textbf{\begin{tabular}[c]{@{}c@{}}In-domain\\ F1 ($\uparrow$)\end{tabular}} & \textbf{\begin{tabular}[c]{@{}c@{}}In-domain\\ FPRD ($\downarrow$)\end{tabular}} & \textbf{\begin{tabular}[c]{@{}c@{}}IPTTS\\ FPRD ($\downarrow$)\end{tabular}} & \textbf{\begin{tabular}[c]{@{}c@{}}NYT\\ Acc ($\uparrow$)\end{tabular}} & \textbf{\begin{tabular}[c]{@{}c@{}}In-domain\\ F1 ($\uparrow$)\end{tabular}} & \textbf{\begin{tabular}[c]{@{}c@{}}In-domain\\ FPRD ($\downarrow$)\end{tabular}} & \textbf{\begin{tabular}[c]{@{}c@{}}IPTTS\\ FPRD ($\downarrow$)\end{tabular}} & \textbf{\begin{tabular}[c]{@{}c@{}}NYT\\ Acc ($\uparrow$)\end{tabular}} & \textbf{\begin{tabular}[c]{@{}c@{}}In-domain\\ Acc. ($\uparrow$)\end{tabular}} & \textbf{\begin{tabular}[c]{@{}c@{}}BROD\\ Acc. ($\uparrow$)\end{tabular}} & \textbf{\begin{tabular}[c]{@{}c@{}}In-domain\\ F1 ($\uparrow$)\end{tabular}} & \textbf{\begin{tabular}[c]{@{}c@{}}Winobias\\ F1-Diff ($\downarrow$)\end{tabular}} \\ \midrule
                           & \multicolumn{4}{c|}{\textit{Non-Transfer (GHC)}}                                                                                                                                                                                                                                                                             & \multicolumn{4}{c|}{\textit{Non-Transfer (Stormfront)}}                                                                                                                                                                                                                                                                              & \multicolumn{2}{c|}{\textit{Non-Transfer (DWMW)}}                                                                                                                                    & \multicolumn{2}{c}{\textit{Non-Transfer (OntoNotes 5.0)}}                                                                                                                                              \\ \midrule
\textbf{Vanilla}           & \textbf{49.60 $\pm$ 1.0}                                                                    & 46.43 $\pm$ 2.5                                                                        & 20.01 $\pm$ 5.7                                                                    & 72.08 $\pm$ 7.3                                                               & \textbf{53.74 $\pm$ 2.8}                                                                    & 18.09 $\pm$ 2.7                                                                        & 11.51 $\pm$ 5.1                                                                    & 73.06 $\pm$ 10                                                                & \textbf{91.46 $\pm$ 0.1}                                                                    & \textbf{78.77 $\pm$ 0.3}                                                                 & \textbf{76.53 $\pm$ 0.2}                                                                    & \textbf{8.04 $\pm$ 0.5}                                                                           \\
\textbf{Emb. Zero}     & 43.76 $\pm$ 0.7                                                                    & 38.31 $\pm$ 2.0                                                                        & 11.95 $\pm$ 2.7                                                                    & \textbf{83.21 $\pm$ 5.2}                                                            & 49.97 $\pm$ 0.6                                                                    & 18.80 $\pm$ 2.0                                                                        & 8.20 $\pm$ 0.3                                                                     & 70.15 $\pm$ 4.4                                                                          & 90.59 $\pm$ 0.1                                                                    & 62.37 $\pm$ 0.4                                                                 & \multicolumn{2}{c}{\cellcolor{na}}                                                                                                                                             \\
\textbf{Expl. Reg.}        & 43.37 $\pm$ 1.8                                                                    & \textbf{29.29 $\pm$ 1.2}                                                                        & \textbf{4.2 $\pm$ 1.6}                                                                      & 81.22 $\pm$ 11            & 51.53 $\pm$ 1.1                                                                    & \textbf{13.43 $\pm$ 1.5}                                                                        & \textbf{3.80 $\pm$ 0.4}                                                                      & \textbf{83.73 $\pm$ 8.0}                                                                                                            & 91.38 $\pm$ 0.1                                                                    & 76.61 $\pm$ 1.5                                                                 & \multicolumn{2}{c}{\cellcolor{na}}                                                                                                                                             \\
\textbf{Adv. Learning}     & \multicolumn{4}{c|}{\cellcolor{na}}                                                                                                                                                                                                                                                                                                   & \multicolumn{4}{c|}{\cellcolor{na}}                                                                                                                                                                                                                                                                                                    & 91.11 $\pm$ 0.3                                                                    & 77.53 $\pm$ 0.9                                                                 & \multicolumn{2}{c}{\cellcolor{na}}                                                                                                                                             \\ \midrule
\textbf{}                  & \multicolumn{4}{c|}{\textit{Stf. $\to$ GHC}}                                                                                                                                                                                                                                                                             & \multicolumn{4}{c|}{\textit{GHC $\to$ Stf.}}                                                                                                                                                                                                                                                                              & \multicolumn{2}{c|}{\textit{FDCL $\to$ DWMW}}                                                                                                            & \multicolumn{2}{c}{\textit{BiasBios $\to$ OntoNotes 5.0}}                                                                                                                  \\ \midrule
\textbf{Van-Transfer}     & 47.83 $\pm$ 2.1                                                                    & 47.51 $\pm$ 4.6                                                                        & 14.00 $\pm$ 0.8                                                                    & 66.71 $\pm$ 10.6                                                              & 55.79 $\pm$ 1.3                                                                    & 17.83 $\pm$ 2.2                                                                        & 8.26 $\pm$ 2.5                                                                     & 76.98 $\pm$ 1.1                                                               & 91.27 $\pm$ 0.2                                                                    & 78.98 $\pm$ 1.1                                                                 & \textbf{76.65 $\pm$ 0.3}                                                                    & 10.54 $\pm$ 0.7                                                                          \\
\textbf{Emb. Zero. Trans.}     & 44.51 $\pm$ 0.5                                                                    & \textbf{40.92 $\pm$ 4.2}                                                                        & 12.91 $\pm$ 0.2                                                                    & \textbf{80.11 $\pm$ 1.2}                                                               & 52.98 $\pm$ 0.6                                                                    & \textbf{16.35 $\pm$ 0.9}                                                                        & 8.04 $\pm$ 2.1                                                                     & 81.11 $\pm$ 1.9                                                               & \textbf{91.53 $\pm$ 0.0}                                                                    & 81.01 $\pm$ 0.9                                                                 & \multicolumn{2}{c}{\cellcolor{na}}                                                                                                                                             \\
\textbf{UBM$_{Reg}$}       & \textbf{49.94 $\pm$ 1.0}                                                                    & 42.71 $\pm$ 3.8                                                                        & \textbf{12.23 $\pm$ 3.3}                                                                    & 75.34 $\pm$ 4.8                                                               & \textbf{56.43 $\pm$ 0.6}                                                                    & 18.03 $\pm$ 2.5                                                                        & \textbf{6.86 $\pm$ 1.1}                                                                     & \textbf{81.18 $\pm$ 1.1}                                                               & 91.39 $\pm$ 0.0                                                                    & 80.27 $\pm$ 0.2                                                                 & \multicolumn{2}{c}{\cellcolor{na}}                                                                                                                                             \\
\textbf{UBM$_{Adv}$}       & \multicolumn{4}{c|}{\cellcolor{na}}                                                                                                                                                                                                                                                                                                   & \multicolumn{4}{c|}{\cellcolor{na}}                                                                                                                                                                                                                                                                                                    & 91.20  $\pm$ 0.0                                                                 & \textbf{81.24 $\pm$ 0.2}                                                                & 76.34 $\pm$ 0.2                                                                    & \textbf{9.27 $\pm$ 1.4}                                                                           \\ \bottomrule
\end{tabular}
}
\caption{\textbf{Cross-domain and task~\method~with a single bias factor}. The source datasets are noted before arrow $(\rightarrow)$. All metrics except In-domain F1 or In-domain Accuracy measures bias. The preferred outcomes for each metric are marked with arrows. The main comparators of~\method~are \textbf{Vanilla}, \textbf{Van-Transfer}, \textbf{Emb. Zero}, and \textbf{Emb. Zero. Trans} that do not perform downstream bias mitigation. We see~\method~maintains in-domain prediction performance while overall reduces bias. Results of Adv. Learning and \method$_{Adv}$ on GHC, Stf. are not included because applying adversarial de-biasing to reduce group identifier bias yields degenerated classifiers.}
\label{tab:cross_full}
\vspace{-0.2cm}
\end{table*}


In this section, we present the results of~\method~in three settings following the order in Sec.~\ref{ssec:rq}: transferring to the same data distribution, transferring to different data distributions, transferring from an upstream model with bias mitigation for multiple bias factors. We follow these main analyses with an investigation of the impact of freezing encoder weights before downstream fine-tuning, and lastly with a brief exploration of how~\method's positive results are achieved. 

\smallskip
\noindent
\textbf{Implementation Details.}
In all experiments reported on below, models are initially fine-tuned from RoBERTa-base. The upstream model is trained for a fixed number of epochs and the checkpoint with the best prediction performance is transferred to the downstream model. See Appendix for more implementation details.   
We use $\mathcal{D}_s \to \mathcal{D}_t $ as the transfer notation, in which upstream and downstream datasets are respectively represented in the left and right-hand side of the arrow. 

\vspace{-0.1cm}
\subsection{\method~with the Same Data Distribution}
\label{ssec:results_same_domain}
\vspace{-0.1cm}

We first briefly show the results when the downstream model sees new, unseen samples from the same data distribution as the upstream model. In this controlled setting, we isolate and test the basic viability of~\method, which requires that information from the upstream model is retained during downstream fine-tuning. 
GHC, Stormfront, FDCL and BiasBios were partitioned into two subsets with equal size, noted as subsets A and B of corresponding datasets, to train the upstream and downstream models respectively. 

Table~\ref{tab:same_domain_partial} presents the results for mitigating group identifier bias in the GHC. 
We see an overall bias reduction, via~\method, by comparing with Vanilla training and Van-Transfer. We include full results and discussions for this simple setting in Appendix.  


\subsection{Cross-domain and Task \method}

Following the result that~\method~is effective in the same-domain setting, we now move to analyzing cross-domain settings in greater depth. 
For hate speech classification, we perform transfer learning from GHC to Stormfront and from Stormfront to GHC; and for toxicity classification, we perform transfer learning from FDCL to DWMW. We also perform transfer learning from BiasBios (occupation prediction) to OntoNotes 5.0 (coreference resolution). 
Table~\ref{tab:cross_full} shows the results of cross-domain and task transfer learning and non-transfer baselines. Our findings are summarized below.

\textbf{\method~can reduce bias in different target domains and tasks compared to fine-tuning without bias mitigation.} The results of cross-domain and task transfer learning (\ie, Stf.$\to$GHC, GHC$\to$Stf., FDCL$\to$DWMW), show that  transferring from a less biased upstream model (\method$_{Reg}$ and \method$_{Adv}$) leads to better downstream bias mitigation compared to directly training without bias mitigation in the target domain (Vanilla). 
Meanwhile, the in-domain classification performance has improved (on GHC and Stormfront) or been preserved (on DWMW). It is notable that directly mitigating bias (Expl. Reg., Adv. Learning) on DWMW is not effective, which is previously observed by~\citet{Xia2020DemotingRB}, while transferring from FDCL is successful. 

There are exceptions where \method~fails to reduce bias. We see the in-domain FPRD on Stormfront does not improve; however, as discussed in our metrics section, the in-domain FPRD is computed over a much smaller set of examples compared to NYT and IPTTS datasets, and is thus less reliable. \method~does not reduce bias compared to Vanilla training on OntoNotes 5.0, but achieves less bias compared to Van-Transfer. This result confirms the effect of bias mitigation in upstream models, but the transfer learning itself has increased the bias. 

\textbf{Comparison with Emb. Zero and Emb. Zero. Trans}. We find two alternative methods, Emb. Zero and Emb. Zero Trans, also reduce bias on some of the datasets. On GHC, Emb. Zero achieves an in-domain FPRD and IPTTS-FPRD lower than \method. However, it comes with clear drop of in-domain classification performance.



\begin{savenotes}
\begin{table*}[t]
\vspace{-0.1cm}
\centering
\scalebox{0.53}{

\begin{tabular}{@{}l|cccc|cc|cc@{}}
\toprule
\textbf{Bias Factor}    & \multicolumn{4}{c|}{\textbf{Group Identifier Bias}}                                                                                                                & \multicolumn{2}{c|}{\textbf{AAVE Dialect Bias}}                                   & \multicolumn{2}{c}{\textbf{Gender Stereotypical Bias}}                                                \\ \midrule
\textbf{Metrics}             & \textbf{In-domain F1 ($\uparrow$)} & \textbf{In-domain FPRD ($\downarrow$)} & \textbf{IPTTS FPRD ($\downarrow$)} & \textbf{NYT Acc ($\uparrow$)} & \textbf{In-domain F1 ($\uparrow$)} & \textbf{BROD Acc. ($\uparrow$)} & \textbf{In-domain F1 ($\uparrow$)} & \textbf{Winobias F1-Diff ($\downarrow$)} \\ \midrule
\textbf{Upstream model}      & \multicolumn{8}{c}{\textit{Stormfront + FDCL}}                                                                                                                                                                                                                                                         \\ \midrule
\textbf{Downstream model}    & \multicolumn{4}{c|}{\textbf{GHC}}                                                                                                                & \multicolumn{2}{c|}{\textbf{DWMW}}                                   & \multicolumn{2}{c}{\cellcolor{na}}                                                \\ \midrule
\textbf{Van-Transfer}        & 49.71 $\pm$ 0.3   \greenarrowup               & \textbf{45.84 $\pm$ 3.8}     \greenarrowdown         & 12.43 $\pm$ 2.5      \greenarrowdown             & \textbf{72.37 $\pm$ 7.4}      \greenarrowup          & 91.64 $\pm$ 0.2      \greenarrowup                    & 81.12 $\pm$ 0.1      \greenarrowup           & \multicolumn{2}{c}{\cellcolor{na}}                                                         \\
\textbf{UBM$_{Reg,Reg}$}     & \textbf{50.21 $\pm$ 1.4}    \greenarrowup          & 47.63 $\pm$ 0.7      \hiddenarrow             & \textbf{12.29 $\pm$ 2.7}      \greenarrowdown           & 68.44 $\pm$ 8.6     \hiddenarrow                 & \textbf{91.66 $\pm$ 0.2}    \greenarrowup                         & 80.05 $\pm$ 0.1        \greenarrowup            & \multicolumn{2}{c}{\cellcolor{na}}                                                         \\
\textbf{UBM$_{Reg,Adv}$}     & 49.89 $\pm$ 1.7     \greenarrowup          & 47.85 $\pm$ 1.2        \hiddenarrow                       & 21.25 $\pm$ 2.0     \hiddenarrow                      & 65.78 $\pm$ 5.7       \hiddenarrow               & 91.55 $\pm$ 0.2      \greenarrowup               & \textbf{81.14 $\pm$ 1.5}      \greenarrowup          & \multicolumn{2}{c}{\cellcolor{na}}                                                         \\ \midrule
\textbf{Upstream Model}      & \multicolumn{8}{c}{\textit{GHC + FDCL}}                                                                                                                                                                                                                                                                \\ \midrule
\textbf{Downstream Model}    & \multicolumn{4}{c|}{\textbf{Stormfront}}                                                                                                         & \multicolumn{2}{c|}{\textbf{DWMW}}                                   & \multicolumn{2}{c}{\textbf{\cellcolor{na}}}                                                \\ \midrule
\textbf{Van-Transfer}        & \textbf{56.78 $\pm$ 1.6}      \greenarrowup        & \textbf{14.26  $\pm$ 0.8}  \greenarrowdown        & 11.04 $\pm$ 0.7   \greenarrowdown            & 77.06 $\pm$ 5.1     \greenarrowup             & 91.65 $\pm$ 0.1      \greenarrowup                 & 80.98 $\pm$ 0.4     \greenarrowup              & \multicolumn{2}{c}{\cellcolor{na}}                                                         \\
\textbf{UBM$_{Reg,Reg}$}     & 53.87 $\pm$ 1.2    \greenarrowup        & 15.92 $\pm$ 1.2     \greenarrowdown                         & \textbf{8.40 $\pm$ 1.4}   \greenarrowdown                     & 83.71 $\pm$ 3.2      \greenarrowup                 & \textbf{91.79 $\pm$ 0.4}          \greenarrowup            & \textbf{81.36 $\pm$ 0.8}     \greenarrowup           & \multicolumn{2}{c}{\cellcolor{na}}                                                         \\
\textbf{UBM$_{Reg,Adv}$}     & 53.63 $\pm$ 0.7  \hiddenarrow                        & 15.52 $\pm$ 2.2     \greenarrowdown                         & 8.90 $\pm$ 1.6   \greenarrowdown                   & \textbf{84.87 $\pm$ 1.1}     \greenarrowup           & 91.33 $\pm$ 0.1       \hiddenarrow                   & 81.09 $\pm$ 0.4       \greenarrowup           & \multicolumn{2}{c}{\cellcolor{na}}                                                         \\ \midrule
\textbf{Upstream Model}      & \multicolumn{8}{c}{\textit{GHC + FDCL + BiasBios}}                                                                                                                                                                                                                                                     \\ \midrule
\textbf{Downstream Model}             & \multicolumn{4}{c|}{\textbf{Stormfront}}                                                                                                         & \multicolumn{2}{c|}{\textbf{DWMW}}                                   & \multicolumn{2}{c}{\textbf{OntoNotes 5.0}}                                    \\ \midrule
\textbf{Van-Transfer}        & \textbf{55.47 $\pm$ 0.7}       \greenarrowup          & \textbf{16.74 $\pm$ 1.5}    \greenarrowdown               & 12.19 $\pm$ 0.7      \greenarrowdown              & 64.15 $\pm$ 6.5     \hiddenarrow           & 91.58 $\pm$ 0.1       \greenarrowup               & 80.74 $\pm$ 0.3      \greenarrowup            &                   73.64 $\pm$ 0.3    \hiddenarrow       &           9.91 $\pm$ 0.2    \hiddenarrow        \\
\textbf{UBM$_{Reg,Reg,Adv}$} & 52.59 $\pm$ 0.5  \hiddenarrow                         & 21.17 $\pm$ 2.0       \hiddenarrow                    & \textbf{9.99 $\pm$ 2.8}     \greenarrowdown                & \textbf{74.58 $\pm$ 4.9}        \greenarrowup          & \textbf{91.64 $\pm$ 0.3}   \greenarrowup                      & 81.07 $\pm$ 0.4      \greenarrowup           & 75.68      \hiddenarrow                    & \textbf{4.93}\footnote{We find~\method$_{Reg, Reg, Adv}$ yield degenerated classifiers for OntoNotes (Test F1 $<46.00$) in 5 out of 6 runs. The result is from one successful run.} \greenarrowdown                             \\
\textbf{UBM$_{Reg,Adv,Adv}$} & 52.85 $\pm$ 0.9       \hiddenarrow              & 18.55 $\pm$ 5.8     \hiddenarrow                 & 13.15 $\pm$ 3.7       \greenarrowdown         & 70.00 $\pm$ 5.8    \hiddenarrow             & 91.50 $\pm$ 0.1    \greenarrowup                  & \textbf{81.08 $\pm$ 0.3}    \greenarrowup        &    \textbf{76.01 $\pm$ 0.4} \hiddenarrow     &   8.67 $\pm$ 0.7    \hiddenarrow          \\ \bottomrule
\end{tabular}
}

\caption{\textbf{Dealing with multiple bias factors with a single upstream model with~\method.} We test three combination of upstream datasets, namely Stormfront + FDCL, GHC + FDCL, and GHC + FDCL + BiasBios, in reducing two or three bias factors. \greenarrowup ~and \greenarrowdown~show whether the metrics has increased or decreased (both imply improvement) compared to non-transfer \textbf{Vanilla} training in Table~\ref{tab:cross_full}.}
\vspace{-0.3cm}
\label{tab:mtl_full}
\end{table*}
\end{savenotes}
\begin{table}[!t]
\centering
\scalebox{0.51}{
\begin{tabular}{@{}l|cccc|cc@{}}
\toprule
\textbf{Metrics}     & \textbf{\begin{tabular}[c]{@{}c@{}}In-domain\\ Acc. ($\uparrow$)\end{tabular}} & \textbf{\begin{tabular}[c]{@{}c@{}}In-domain\\ FPRD ($\downarrow$)\end{tabular}} & \textbf{\begin{tabular}[c]{@{}c@{}}IPTTS\\ FPRD ($\downarrow$)\end{tabular}} & \textbf{\begin{tabular}[c]{@{}c@{}}NYT\\ Acc. ($\uparrow$)\end{tabular}} & \textbf{\begin{tabular}[c]{@{}c@{}}In-domain\\ Acc. ($\uparrow$)\end{tabular}} & \textbf{\begin{tabular}[c]{@{}c@{}}BROD\\ Acc. ($\uparrow$)\end{tabular}} \\ \midrule
                     & \multicolumn{4}{c|}{\textit{Stf. $\to$  GHC}, \method$_{Reg}$}                                                                                                                                                                                                                                                                                & \multicolumn{2}{c}{\textit{FDCL $\to$ DWMW}, \method$_{Adv}$}                                                                                                              \\ \midrule
Freeze               & 45.42                                                                          & \textbf{37.71}                                                                            & \textbf{7.82}                                                                          & \textbf{84.45}                                                                    & 83.25                                                                          & 64.80                                                                     \\
\textbf{$\ell^2$-sp} & 49.31                                                                          & 47.03                                                                            & 14.24                                                                         & 71.88                                                                    & \textbf{91.38}                                                                          & 79.95                                                                     \\
\textbf{Fine-tune}   & \textbf{49.94}                                                                          & 42.71                                                                            & 12.23                                                                         & 75.34                                                                    & 91.20                                                                          & \textbf{81.24}                                                                     \\ \midrule
\textbf{}            & \multicolumn{4}{c|}{\textit{GHC $\to$ Stf.} \method$_{Reg}$}                                                                                                                                                                                                                                                                                 & \multicolumn{2}{c}{\cellcolor{na}}                                                                                                                                     \\ \midrule
\textbf{Freeze}      & 47.32                                                                          & 25.02                                                                            & 8.24                                                                          & 64.60                                                                    & \multicolumn{2}{c}{\cellcolor{na}}                                                                                                                                     \\
\textbf{$\ell^2$-sp} & 55.80                                                                          & 19.75                                                                            & 6.72                                                                          & 80.42                                                                    & \multicolumn{2}{c}{\cellcolor{na}}                                                                                                                                     \\
\textbf{Fine-tune}   & \textbf{56.43}                                                                          & 18.03                                                                            & 6.86                                                                          & \textbf{81.18}                                                                    & \multicolumn{2}{c}{\cellcolor{na}}                                                                                                                                     \\ \bottomrule
\end{tabular}}
\vspace{-0.1cm}
\caption{\method~while keeping the encoder frozen (Freeze), discouraging parameter changes ($\ell^2$-sp), or standard fine-tuning (Fine-tune). We see weight freezing and $\ell^2$-sp overall do not improve over simple fine-tuning on Stf. $\to$ GHC and FDCL $\to$ DWMW.}
\label{tab:frz_lsp}
\end{table}

\vspace{-0.1cm}
\subsection{Mitigating Multiple Bias Factors}
\vspace{-0.1cm}

Having observed an overall positive effect of~\method~across domains and tasks, next we present the results of experiments on mitigating multiple bias factors with a single upstream model. This involves training an upstream model with multiple bias mitigation objectives across multiple datasets, followed by fine-tuning on a single dataset without bias mitigation.
We test three combinations of datasets. First, a multi-task model is trained to jointly mitigate group identifier bias and AAVE dialect bias using GHC and FDCL (GHC + FDCL), and transferred to Stormfront and DWMW. Next, a model is similarly trained jointly on group identifier and AAVE biases on and Stormfront and FDCL (Stf. + FDCL) and transferred to GHC and DWMW. Lastly, models were trained over source datasets GHC, FDCL, BiasBios (GHC+FDCL+BiasBios) to mitigate all three bias factors, and transferred to Stormfront, DWMW, and OntoNotes. 
The results are shown in Table~\ref{tab:mtl_full}.

\textbf{Comparison to Single-Dataset Vanilla Baselines.} As a basic measure of bias mitigation success, we compare multi-dataset models' results with single-dataset Vanilla training and Van-Transfer. We see \method~with GHC + FDCL successfully reduces both group identifier bias and AAVE dialect bias in downstream models. \method~with GHC + FDCL + BiasBios also successfully reduces group identifier bias in terms of IPTTS, FPRD (which is the most reliable metrics of bias given its large size), and AAVE bias. It also reduces gender stereotypical bias compared to Van-Transfer in some experimental runs, but in an unstable manner, demonstrated by the large variance of F1-Diff and degenerated runs of \method$_{Reg,Reg,Adv}$. 

Results of \method~on Stf. + FDCL are less promising. We find \method$_{Reg,Adv,Adv}$ is not successful in reducing group identifier bias. \method$_{Reg,Reg,Adv}$ could reduce bias on IPTTS-FPRD, but does not improve other metrics. Notably, \method~on Stf. + FDCL clearly underperform~\method~on Stf. only. 

\textbf{\method$_{Reg}$ versus \method$_{Adv}$.} Empirically, we find using explanation regularization on FDCL (\method$_{reg,reg}$, \method$_{reg,reg,adv}$) instead of adversarial learning (\method$_{reg,adv}$, \method$_{reg,adv,adv}$) consistently improves bias mitigation performance on other bias factors.

\textbf{Takeaways.} Our results show it is possible to reduce multiple bias factors via \method. However, we have shown that these effects are not automatic for each new dataset added to upstream models for multi-task bias mitigation. 



\vspace{-0.1cm}
\subsection{Freezing or Regularizing Model Weights}
\label{ssec:frz_unfrz}
\vspace{-0.1cm}

In the experiments above, we have shown that the effect of mitigating bias is partially preserved with simple fine-tuning. Next, we study whether freezing the encoders or discouraging their weight changes improves bias mitigation in the target domain, as they intuitively try to retain effect of bias mitigation.
However, we find a counter-intuitive result: these approaches typically do not achieve reduced downstream bias, and in fact reduce in-domain classification performance. Table~\ref{tab:frz_lsp} shows the results when we keep the weights frozen (Freeze), discouraging weights from changing with $\ell^2$-sp regularizer~\citep[][details in appendix]{Li2018ExplicitIB}, or standard fine-tuning (fine-tune). In Stf. $\to$ GHC, freezing the weights contributed to reducing the bias, while $\ell^2$-sp failed to help. 
In GHC $\to$ Stf and FDCL $\to$ DWMW, freezing the weights and $\ell^2$-sp both increased the bias. 
A possible reason is that by freezing the encoder, we reduce its expressive power. As a result, the encoder is prone to capture simple but spurious correlations. 

\subsection{Investigating Why UBM Reduces Bias}
\label{ssec:interp}
\vspace{-0.1cm}
\begin{figure}
    \centering
    \vspace{-0.3cm}
    \subfloat[][Gab A $\to$ Gab B]{\includegraphics[width=0.20\textwidth]{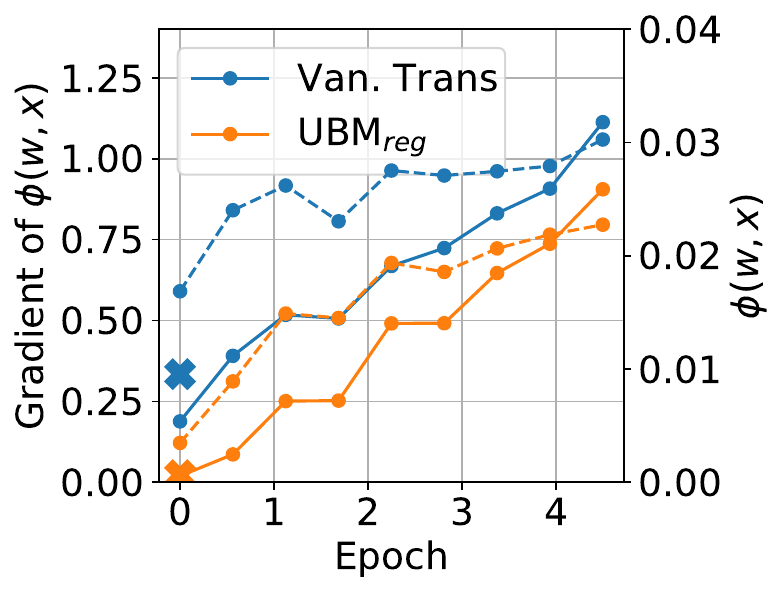}}
    \subfloat[][Stf. A $\to$ Stf. B]{\includegraphics[width=0.20\textwidth]{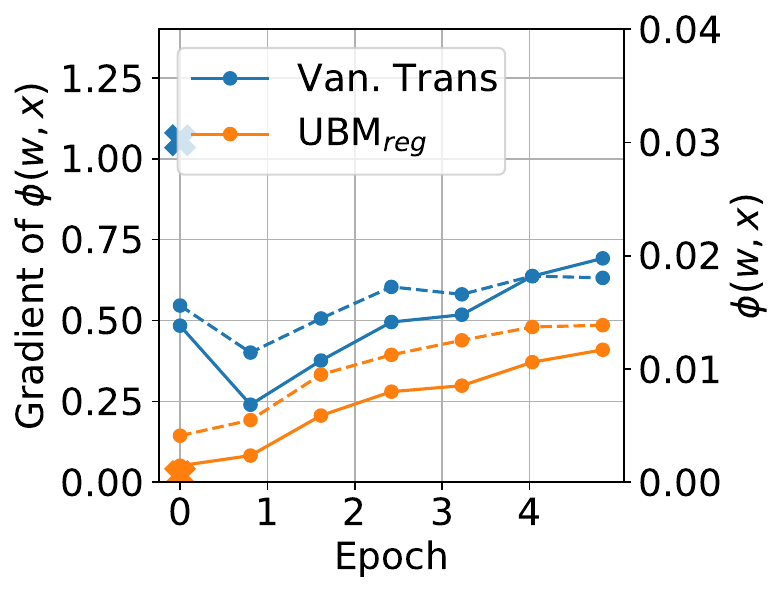}}
    
    \subfloat[][Gab $\to$ Stf.]{\includegraphics[width=0.20\textwidth]{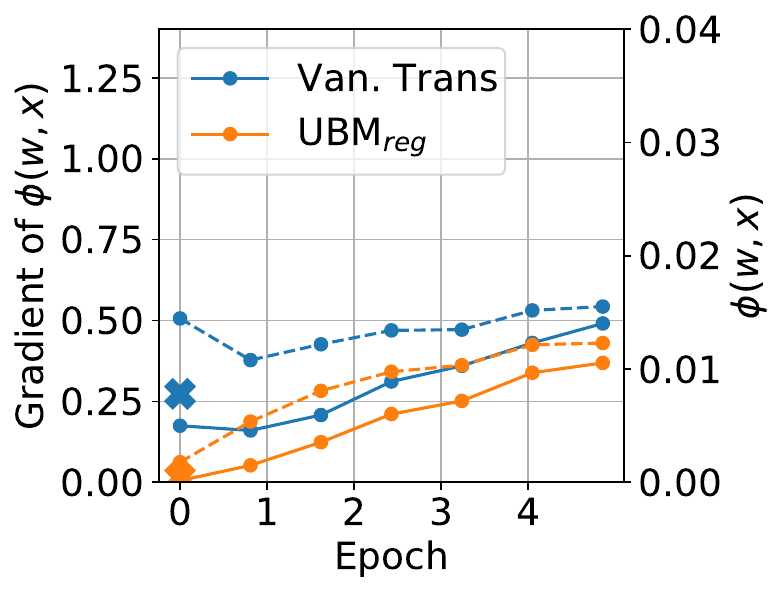}}
    \subfloat[][Stf. $\to$ Gab]{\includegraphics[width=0.20\textwidth]{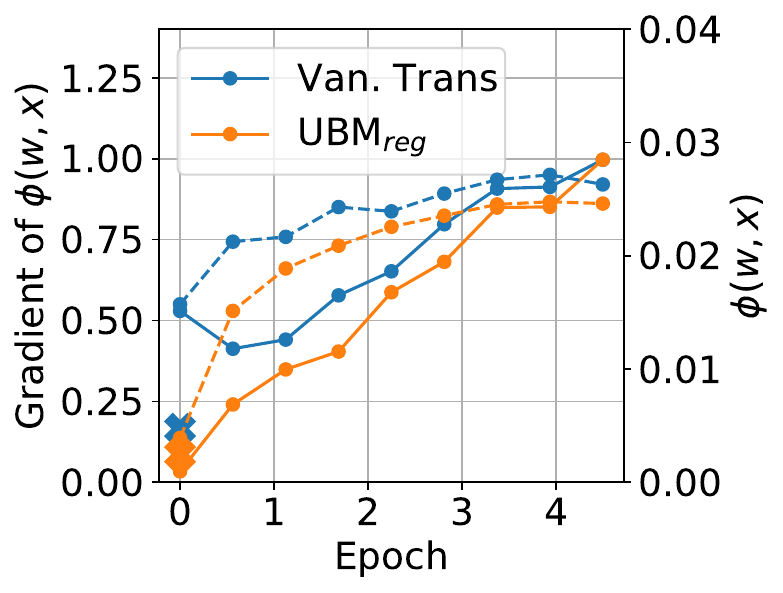}}
    \vspace{-0.1cm}
    \caption{Gradient of importance attribution on group identifiers $\phi(w,
    \rvx)$ over time (in solid lines) and the corresponding values of $\phi(w,\rvx)$ (in dash lines) during downstream fine-tuning. The cross-marks show the gradient measured in the upstream model (before re-initialization of the classifier layer). \method$_{reg}$ not only reduces importance attributed to group identifiers, but also the gradient norm of the importance. 
    }
    \label{fig:analysis}
\vspace{-0.2cm}
\end{figure}

We attempt to interpret why fine-tuning from a de-biased upstream model remains less biased during fine-tuning from the perspective of gradient of importance attributed to words $w$ related to bias factors (\eg, group identifiers) by the input occlusion algorithm. A large importance attribution usually induces bias. Figure~\ref{fig:analysis} plots the importance attribution of group identifiers $\phi(w,\rvx)$ and the norm of its gradient w.r.t. parameters $\theta$ of the encoder $g$, noted as $||\nabla_{\theta} \phi(w, \rvx)||_2$.



\textbf{\method~reduces the gradient of $\phi(w, \rvx)$, so that $\phi(w, \rvx) $ is less likely to change \textit{at the beginning} of downstream fine-tuning}. Fig.~\ref{fig:analysis} shows~\method~has not only reduced value of importance attributed to spurious patterns, but also reduced their gradients. 
The gradient norm is highly indicative about how the importance $\phi(w, \rvx)$ will change in the downstream model, because when the loss in Eq.~\ref{eq:expl_reg} in the upstream model is minimized, the gradient $\nabla_{\theta} \phi(w, \rvx)$ has the same norm but the opposite direction as the main downstream classification objective $\nabla_\theta \ell_c$. It implies that whether the upstream model converges at an optimum where both objectives agree (\ie, gradients are small) can be an important indicator of the success of \method.

The figure further shows that the gradient and the value of $\phi(w, \rvx)$ remain small for \method$_{reg}$ over the whole training process. We leave more study into the training dynamics of~\method~as future works.

\section{Related Works}
\label{sec:related}
\vspace{-0.1cm}


Here we review approaches that inform the present work (techniques for bias mitigation) and are related to the basic idea of \method.

\smallskip
\noindent
\textbf{Mitigating bias in representations.} Bias can be mitigated directly in representations of data.~\citet{Zhang2018MitigatingUB, Beutel2017DataDA} proposed training a classifier together with an adversarial predictor for sensitive attributes.~\citet{Madras2018LearningAF} further studied re-usable de-biased representations by training a new downstream classifier (potentially with a different classification task) using the learned representations.  However, this practice relies on frozen representations (rather than models themselves), which precludes the possibility of generating predictions for new data.

\smallskip
\noindent
\textbf{Mitigating bias in pretrained models.} Another line of work addresses bias in pretrained models~\citep[\eg, word vectors, BERT,][]{Zhou2019ExaminingGB, May2019OnMS, Bhardwaj2020InvestigatingGB, Liang2020TowardsDS}. Many such studies again focus on bias in frozen data representations, and do not study their effects on downstream classifiers. 
Others alternatively assess the propagation of bias from pretrained models to downstream classifiers:
~\citet{Ravfogel2020NullIO} study algorithms for mitigating bias in pretrained models by de-biasing the learned representations, which can subsequently be used in classifiers as frozen representations. 


\smallskip
\noindent
\textbf{Transferring learning of fairness and robustness.} 
A few previous works have studied related research problems, with significant differences to our work. Though~\citet{Schumann2019TransferOM} theoretically analyzes the transferability of fairness across domains, it assumes simultaneous access of source and target domain data, which does not account for transferring upstream bias mitigation to arbitrary downstream fine-tuned models.~\citet{Shafahi2020AdversariallyRT} study transfer learning of robustness to adversarial attacks under fine-tuning, but do not seek to mitigate bias.

\vspace{-0.1cm}
\vspace{-0.1cm}
\section{Conclusion}
\vspace{-0.1cm}

We observe that the effects of bias mitigation are indeed transferable in fine-tuning LMs. Future works in fine-tuning LMs can use~\method~ in order to easily apply the positive effects of bias mitigation methods to new domains and tasks without customized bias mitigation processes or access to sensitive user information. Though~\method~does not rival directly mitigating bias on the downstream task, it is more efficient and accessible. Future works can develop the effectiveness of~\method~beyond the default scenarios in this paper, and potentially apply it to tasks and settings beyond hate speech, toxicity classification, occupation prediction, and coreference resolution in English corpora. 

\subsection*{Broader Impact Statement}

Our analysis demonstrates the effectiveness of~\methodlong. As we stated in the paper, the reduced efforts of downstream bias mitigation will facilitate broader application of bias mitigation in the growing deep learning community.

While we may expect to obtain an ``off-the-shelf'' language model that could reduce multiple kinds of bias with~\method, we emphasize that proper evaluation of bias may still be required in downstream side, especially for guaranteed bias mitigation. Currently, our initial analysis of~\method~confirms that bias mitigation effects are transferable, but does not provide guarantees of bias mitigation or levels of bias mitigation in the direct setting. The findings in this analysis should identify the potential of~\method~to the broader NLP and machine learning communities, which may be extended with new approaches within the \method~framework, or interpretation techniques (as in Sec.~\ref{ssec:interp}).

\bibliography{fair}
\bibliographystyle{acl_natbib}

\clearpage
\appendix
\begin{savenotes}
\begin{table*}[]
\centering
\scalebox{0.53}{

\begin{tabular}{@{}l|cccc|cccc|cc|cc@{}}
\toprule
\textbf{Method / Datasets} & \multicolumn{4}{c|}{\textbf{GHC B}}                                                                                                                                                                                                                                                                                      & \multicolumn{4}{c|}{\textbf{Stormfront B}}                                                                                                                                                                                                                                                                                & \multicolumn{2}{c|}{\textbf{FDCL B}}                                                                                                                       & \multicolumn{2}{c}{\textbf{Biasbios B}}                                                                                                                           \\ \midrule
\textbf{Metrics}           & \textbf{\begin{tabular}[c]{@{}c@{}}In-domain\\ F1 ($\uparrow$)\end{tabular}} & \textbf{\begin{tabular}[c]{@{}c@{}}In-domain\\ FPRD ($\downarrow$)\end{tabular}} & \textbf{\begin{tabular}[c]{@{}c@{}}IPTTS\\ FPRD ($\downarrow$)\end{tabular}} & \textbf{\begin{tabular}[c]{@{}c@{}}NYT\\ Acc ($\uparrow$)\end{tabular}} & \textbf{\begin{tabular}[c]{@{}c@{}}In-domain\\ F1 ($\uparrow$)\end{tabular}} & \textbf{\begin{tabular}[c]{@{}c@{}}In-domain\\ FPRD ($\downarrow$)\end{tabular}} & \textbf{\begin{tabular}[c]{@{}c@{}}IPTTS\\ FPRD ($\downarrow$)\end{tabular}} & \textbf{\begin{tabular}[c]{@{}c@{}}NYT\\ Acc ($\uparrow$)\end{tabular}} & \textbf{\begin{tabular}[c]{@{}c@{}}In-domain\\ Acc. ($\uparrow$)\end{tabular}} & \textbf{\begin{tabular}[c]{@{}c@{}}BROD\\ Acc. ($\uparrow$)\end{tabular}} & \textbf{\begin{tabular}[c]{@{}c@{}}In-domain\\ F1 ($\uparrow$)\end{tabular}} & \textbf{\begin{tabular}[c]{@{}c@{}}Winobias\\ F1-Diff ($\downarrow$)\end{tabular}} \\ \midrule
                           & \multicolumn{4}{c|}{\textit{Non-Transfer (GHC B)}}                                                                                                                                                                                                                                                                       & \multicolumn{4}{c|}{\textit{Non-Transfer (Stf. B)}}                                                                                                                                                                                                                                                                       & \multicolumn{2}{c|}{\textit{Non-Transfer (FDCL. B)}}                                                                                                       & \multicolumn{2}{c}{\textit{Non-Transfer (Biasbios B)}}                                                                                                            \\ \midrule
\textbf{Vanilla}           & 37.91 $\pm$ 2.5                                                                    & 35.64 $\pm$ 2.2                                                                        & 21.50 $\pm$ 2.8                                                                    & 68.55 $\pm$ 20                                                                & \textbf{55.56 $\pm$ 0.5}                                                                    & 20.81 $\pm$ 4.9                                                                        & 10.99 $\pm$ 5.6                                                                    & \textbf{66.28 $\pm$ 8.1}                                                               & 75.72 $\pm$ 0.2                                                                      & 73.57 $\pm$ 1.2                                                                 & \textbf{85.52 $\pm$ 0.1}                                                                    & 13.83 $\pm$ 0.2                                                                          \\
\textbf{Expl. Reg.}        & \textbf{38.09 $\pm$ 2.7}                                                                    & \textbf{18.68 $\pm$ 0.3}                                                                        & \textbf{4.82 $\pm$ 1.1}                                                                     & \textbf{84.05 $\pm$ 3.0}                                                               & 53.05 $\pm$ 1.0                                                                    & \textbf{15.97 $\pm$ 1.1}                                                                       & \textbf{3.36 $\pm$ 3.3}                                                                     & 65.23 $\pm$ 10                                                                & \textbf{77.30 $\pm$ 0.2}                                                                      & 76.72 $\pm$ 1.1                                                                 & \multicolumn{2}{c}{\cellcolor{na}}                                                                                                                                             \\
\textbf{Adv. Learning}     & \multicolumn{4}{c|}{\cellcolor{na}}                                                                                                                                                                                                                                                                                                   & \multicolumn{4}{c|}{\cellcolor{na}}                                                                                                                                                                                                                                                                                                    & 75.28 $\pm$ 0.2                                                                      & \textbf{77.12 $\pm$ 1.2}                                                                 & 85.07 $\pm$ 0.0                                                                    & \textbf{9.61 $\pm$ 0.5}                                                                           \\ \midrule
\textbf{}                  & \multicolumn{4}{c|}{\textit{GHC A $\to$ GHC B}}                                                                                                                                                                                                                                                                          & \multicolumn{4}{c|}{\textit{Stf. A $\to$ Stf. B}}                                                                                                                                                                                                                                                                         & \multicolumn{2}{c|}{\textit{FDCL A $\to$ FDCL B}}                                                                                                          & \multicolumn{2}{c}{\textit{Biasbios A $\to$ Biasbios B}}                                                                                                          \\ \midrule
\textbf{Van. Transfer}     & 42.41 $\pm$ 1.0                                                                    & 37.44 $\pm$ 1.5                                                                        & 17.67 $\pm$ 2.1                                                                    & 75.35 $\pm$ 4.2                                                               & 58.43 $\pm$ 1.2                                                                    & 17.96 $\pm$ 3.6                                                                        & 11.58 $\pm$ 4.8                                                                    & \textbf{74.12 $\pm$ 4.2}                                                               & \textbf{76.33 $\pm$ 0.6}                                                                      & 70.35 $\pm$ 2.4                                                                 & 85.81 $\pm$ 0.2                                                                      &  12.59 $\pm$ 0.5                                                        \\
\textbf{UBM$_{Reg}$}       & \textbf{43.79 $\pm$ 1.9}                                                                    & \textbf{34.34 $\pm$ 3.1}                                                                        & \textbf{10.02 $\pm$ 1.1}                                                                    & \textbf{81.40 $\pm$ 1.4}                                                               & \textbf{58.56 $\pm$ 1.0}                                                                    & \textbf{16.42 $\pm$ 0.9}                                                                        & \textbf{7.51 $\pm$ 2.4}                                                                     & 69.45 $\pm$ 4.0                                                               & 76.22 $\pm$ 0.5                                                                      & 69.29 $\pm$ 1.8                                                                 & \multicolumn{2}{c}{\cellcolor{na}}                                                                                                                                             \\
\textbf{UBM$_{Adv}$}       & \multicolumn{4}{c|}{\cellcolor{na}}                                                                                                                                                                                                                                                                                                   & \multicolumn{4}{c|}{\cellcolor{na}}                                                                                                                                                                                                                                                                                                    & 75.88 $\pm$ 0.4                                                                      & \textbf{71.11 $\pm$ 1.6}                                                                 & \textbf{85.86 $\pm$ 0.1}                                                             & \textbf{11.99 $\pm$ 0.5}                                                                              \\ \bottomrule
\end{tabular}

}

\caption{\textbf{Same-domain and task UBM with a single bias factor.} We partition a dataset to two subsets, noted as split A and split B. We train the upstream model with split A and fine-tune on split B.  All metrics except In-domain F1 or In-domain Accuracy measures bias.  The preferred outcomes for eachmetric are marked with arrows. }
\label{tab:same_domain_full}
\end{table*}
\vspace{0.2cm}
\end{savenotes}
\section{Implementation Details}

\subsection{Training Details}
We use RoBERTa-base as our base model. In the bias mitigation phase, models for GHC, Stormfront, FDCL, DWMW, and BiasBios are trained with a learning rate $1e^{-5}$, and the checkpoint with the best validation F1 or accuracy score is provided to the fine-tuning phase. We train on GHC, FDCL, DWMW, BiasBios for maximum 5 epochs and Stormfront for maximum 10 epochs . The checkpoint with the best validation in-domain classification performance is kept. In the fine-tuning phase, we try the learning rate $1e^{-5}$ and $5e^{-6}$, and report the results with a higher validation in-domain classification performance. For the coreference resolution model on OntoNotes 5.0, we adapt existing code implementation\footnote{\url{https://github.com/mandarjoshi90/coref}}~\cite{joshi2019coref} to support loading RoBERTa-base as the base model. We use the same hyperparameter settings as BERT-base in the provided code implementation. 

To report mean and standard deviation of performance are computed over 3 runs for most of the experiments, with the same set of random seeds; for GHC and Stf. experiments in Table~\ref{tab:cross_full}, and \method$_{reg,reg,adv}$ on OntoNotes 5.0, we run experiments for 6 runs. Models except coreference resolution models on OntoNotes, are trained on a single GTX 2080 Ti GPU. Coreference resolution models are trained on a single Quadro RTX 6000 GPU.

The training time per iteration is consistent over experiments in about 1.5 iteration per second, except the conference resolution. The training of coreference resolution model on OntoNotes 5.0 takes around 8 hours. The largest dataset among other datasets, BiasBios, takes 2 hours to train.

\subsection{Details of Bias Mitigation Algorithms}
For explanation regularization algorithm, we set the regularization strength $\alpha$ as 0.03 for GHC and Stormfront experiments, and 0.1 for FDCL and DWMW experiments. We regularize importance score on 25 group identifiers in~\cite{kennedy2018gab} for GHC and Stormfront. These group identifiers the ones that have the largest coefficient in a bag-of-words linear classifier. For FDCL, we extract 50 words with largest coefficient in the bag-of-words linear classifier with a AAE dialect probability higher than 60\% (given by the off-the-shelf AAE dialect predictor~\cite{Blodgett2016DemographicDV}) on its own. For adversarial de-biasing, the adversarial loss term has the same weight as the classification loss term. 
\begin{savenotes}
\begin{table*}[]
\centering
\scalebox{0.53}{
\vspace{0.1cm}
\begin{tabular}{@{}l|cccc|cc|cc@{}}
\toprule
\textbf{Metrics}             & \textbf{In-domain F1 ($\uparrow$)} & \textbf{In-domain FPRD ($\downarrow$)} & \textbf{IPTTS FPRD ($\downarrow$)} & \textbf{NYT Acc ($\uparrow$)} & \textbf{In-domain F1 ($\uparrow$)} & \textbf{BROD Acc. ($\uparrow$)} & \textbf{In-domain F1 ($\uparrow$)} & \textbf{In-domain TPRD ($\downarrow$)} \\ \midrule
\textbf{Upstream model}      & \multicolumn{8}{c}{\textit{Stormfront A + FDCL A}}                                                                                                                                                                                                                                         \\ \midrule
\textbf{Downstream model}    & \multicolumn{4}{c|}{\textbf{Stormfront B}}                                                                                                       & \multicolumn{2}{c|}{\textbf{FDCL B}}                                 & \multicolumn{2}{c}{\textbf{\cellcolor{na}}}                                    \\ \midrule
\textbf{Van-Transfer}        & \textbf{57.58 $\pm$ 2.7} \greenarrowup                       & \textbf{13.97 $\pm$ 2.0} \greenarrowdown                           & 11.33 $\pm$ 2.2 \hiddenarrow                       & 75.72 $\pm$ 7.5 \greenarrowup                  & \textbf{77.18 $\pm$ 0.5} \greenarrowup                       & 71.12 $\pm$ 1.2 \hiddenarrow                    & \multicolumn{2}{c}{\cellcolor{na}}                                             \\
\textbf{UBM$_{Reg,Reg}$}     & 56.72 $\pm$ 1.7 \greenarrowup                       & 17.91 $\pm$ 1.0 \greenarrowdown                           & \textbf{8.05 $\pm$ 0.6} \greenarrowdown                        & \textbf{77.40 $\pm$ 0.3} \greenarrowup                  & 77.13 $\pm$ 0.3 \greenarrowup                       & 72.17 $\pm$ 1.6 \hiddenarrow                    & \multicolumn{2}{c}{\cellcolor{na}}                                             \\
\textbf{UBM$_{Reg,Adv}$}     & 55.63 $\pm$ 2.5 \greenarrowup                       & 17.14 $\pm$ 0.5 \greenarrowdown                           & 13.78 $\pm$ 4.3 \hiddenarrow                       & 70.37 $\pm$ 10 \greenarrowup                   & 76.64 $\pm$ 0.6 \greenarrowup                       & \textbf{76.55 $\pm$ 0.6} \greenarrowup                    & \multicolumn{2}{c}{\cellcolor{na}}                                             \\ \midrule
\textbf{Upstream Model}      & \multicolumn{8}{c}{\textit{GHC A + FDCL A}}                                                                                                                                                                                                                                                \\ \midrule
\textbf{Downstream Model}    & \multicolumn{4}{c|}{\textbf{GHC B}}                                                                                                              & \multicolumn{2}{c|}{\textbf{FDCL B}}                                 & \multicolumn{2}{c}{\textbf{\cellcolor{na}}}                                    \\ \midrule
\textbf{Van-Transfer}        & \textbf{44.30 $\pm$ 0.7} \greenarrowup                       & 41.06 $\pm$ 3.9 \hiddenarrow                           & 19.75 $\pm$ 6.9 \greenarrowdown                       & 74.60 $\pm$ 6.3 \greenarrowup                  & \textbf{77.34 $\pm$ 0.4} \greenarrowup                       & 72.96 $\pm$ 1.5 \hiddenarrow                    & \multicolumn{2}{c}{\cellcolor{na}}                                             \\
\textbf{UBM$_{Reg,Reg}$}     & 42.96 $\pm$ 2.0 \greenarrowup                       & 33.98 $\pm$ 3.0 \greenarrowdown                           & \textbf{9.30 $\pm$ 2.1} \greenarrowdown                        & \textbf{86.05 $\pm$ 1.9} \greenarrowup                  & 76.21 $\pm$ 0.4 \greenarrowup                       & 73.10 $\pm$ 1.4 \hiddenarrow                    & \multicolumn{2}{c}{\cellcolor{na}}                                             \\
\textbf{UBM$_{Reg,Adv}$}     & 42.44 $\pm$ 3.5 \greenarrowup                       & \textbf{33.96 $\pm$ 1.5} \greenarrowdown                           & 16.68 $\pm$ 1.7 \greenarrowdown                       & 81.79 $\pm$ 9.0 \greenarrowup                  & 76.94 $\pm$ 0.4 \greenarrowup                       & \textbf{76.55 $\pm$ 0.7} \greenarrowup                    & \multicolumn{2}{c}{\cellcolor{na}}                                             \\ \midrule
\textbf{Upstream Model}      & \multicolumn{8}{c}{\textit{GHC A+ FDCL A+ BiasBios A}}                                                                                                                                                                                                                                     \\ \midrule
\textbf{Downstream Model}             & \multicolumn{4}{c|}{\textbf{GHC B}}                                                                                                         & \multicolumn{2}{c|}{\textbf{FDCL B}}                                   & \multicolumn{2}{c}{\textbf{BiasBios B}}                           \\ \midrule
\textbf{Van-Transfer}        &                 42.80 $\pm$ 3.3 \greenarrowup             &           37.83 $\pm$ 10  \hiddenarrow           &         17.38 $\pm$ 1.7   \greenarrowdown             &               72.23 $\pm$ 13  \greenarrowup          &          \textbf{77.30  $\pm$ 0.2}   \greenarrowup       &      72.91 $\pm$ 1.1    \hiddenarrow               &              \textbf{85.81 $\pm$ 0.0}   \greenarrowup     &             12.78 $\pm$ 0.0  \greenarrowdown      \\
\textbf{UBM$_{Reg,Reg,Adv}$} &         \textbf{42.81 $\pm$ 1.9}  \greenarrowup                 &               \textbf{31.86 $\pm$ 1.1}     \greenarrowdown           &               \textbf{9.61 $\pm$ 1.8}    \greenarrowdown          &               82.15 $\pm$ 9.5   \greenarrowup             &         76.95 $\pm$ 0.1     \greenarrowup       &               72.93 $\pm$ 0.6    \hiddenarrow       &              \textbf{85.81 $\pm$ 0.0}     \greenarrowup                  &      \textbf{11.72 $\pm$ 0.8} \greenarrowdown                  \\
\textbf{UBM$_{Reg,Adv,Adv}$} &      41.66 $\pm$ 2.3  \greenarrowup           &             33.39 $\pm$ 0.2 \greenarrowdown           &         10.00 $\pm$ 0.8  \greenarrowdown        &       \textbf{83.50 $\pm$ 3.0}    \greenarrowup         &               77.03 $\pm$    0.3  \greenarrowup      &     \textbf{75.87 $\pm$ 0.8}   \greenarrowup                         &            85.79 $\pm$ 0.1   \greenarrowup         &            12.43 $\pm$ 0.5  \greenarrowdown      \\ \bottomrule
\end{tabular}
}

\caption{\textbf{Dealing with multiple bias factors with a single upstream model with~\method, where the domains and tasks are the same in the upstream and the downstream model}.  \greenarrowup ~and \greenarrowdown~show whether the metrics has increased or decreased (both imply improvement) compared to non-transfer \textbf{Vanilla} training in Table~\ref{tab:cross_full}.}
\label{tab:mtl_same_full}
\vspace{0.2cm}
\end{table*}
\end{savenotes}

\subsection{Dataset Details}
\noindent \textbf{Group Identifier Bias Experiments.} We use a balanced split of the GHC dataset, where training, validation, and the test set consist of 22,767, 1,586, and 1,344 examples. We use the union of ``human degration'' and ``call for violence'' has the hate label, which results in around 9\% of hate examples for all the splits. Note that the split is different from~\citet{Kennedy2020ContextualizingHS} where the test split has a much higher ratio of hate examples. We use the same split as~\citet{Kennedy2020ContextualizingHS} for the Stormfront dataset, with 7,896, 978, and 1,998 examples in training, validation, and test sets. The NYT corpus contain $12.5k$ non-hate sentences for testing.

\noindent \textbf{AAVE Dialect Bias Experiments.} We follow~\citet{Sap2019TheRO} for the split ratio (73/12/15) of the DWMW dataset, which results in 17,994, 2,974, and 3,718 examples in each split. For the FDCL dataset, as only tweet ids are provided and some of the tweets are no longer available, the final dataset consists of 41191, 5149, 5149 examples for each split. We release the tweet ids used in each split. Following~\cite{Xia2020DemotingRB}, we sample 20$k$ examples with an AAVE speaker probability (which is included in the dataset) greater than 80\%. We manually verify a subset of examples in BROD following the protocols of~\cite{Sap2019TheRO} and found ~93\% of sentences clearly non-toxic.

\noindent \textbf{Gender Stereotypical Bias Experiments.} For the BiasBios dataset, we use the same split as~\citet{Ravfogel2020NullIO} with 255,710 training examples (65\%), 39,359 validation examples (10\%), and 98,344 (25\%) test examples. We use the official dataset split~\cite{weischedel2013ontonotes} for the OntoNotes 5.0 dataset.

We use the same train/test splits between “transfer” and “non-transfer” setup. Two partitions of datasets used for Same-distribution ~\method experiments have a random half of total examples for train/validation/test splits. For IPTTS/NYT/BROD, we use the same test set across tables.

\subsection{Details of $\ell^2$-sp Regularizer }
The $\ell^2$-sp regularizer~\citep{Li2018ExplicitIB} we applied in Sec.~\ref{ssec:frz_unfrz} penalizes the distance between the weights and the initial point of fine-tuning. Formally, let $\rvw_0$ be the initial weight of the encoder $g_t$ before fine-tuning, and $\rvw$ be the current weight of $g_t$. The $\ell^2$-sp regularizer is written as $\Omega(\rvw) = \beta || \rvw - \rvw_0 ||^2_2$, appended to the learning objective. $\beta$ is a hyperparameter controlling the strength of the regularization. We reported results where $\beta=1$. We tried different values of $\beta$ from $1e^{-6}$ to 100, increasing $\beta$ by 10 times each time, but we do not see changes in the conclusion.

\section{Complete Analysis of \method~over the Same Data Distribution}
\begin{table}[t]
\vspace{-0.0cm}
\centering
\scalebox{0.62}{
\begin{tabular}{@{}l|cccc@{}}
\toprule
\textbf{Metrics}           & \textbf{\begin{tabular}[c]{@{}c@{}}In-domain\\ F1 ($\uparrow$)\end{tabular}} & \textbf{\begin{tabular}[c]{@{}c@{}}In-domain\\ FPRD ($\downarrow$)\end{tabular}} & \textbf{\begin{tabular}[c]{@{}c@{}}IPTTS\\ FPRD ($\downarrow$)\end{tabular}} & \textbf{\begin{tabular}[c]{@{}c@{}}NYT\\ Acc ($\uparrow$)\end{tabular}}  \\ \midrule
                           & \multicolumn{4}{c}{\textit{Stormfront $\to$ GHC}}                                                                                                                                                                                                                   \\ \midrule
\textbf{Expl. Reg.}           & 43.37 $\pm$ 1.8                                                                &  29.29     $\pm$  1.2                                  &  4.20 $\pm$ 1.6                                                                                                          & 81.22 $\pm$ 11                                                        \\
\textbf{Van-Transfer + Reg.}        & \textbf{45.25 $\pm$ 2.2}                                                              & 29.91 $\pm$ 1.9                                                                       & 5.01 $\pm$ 1.3                                              & 86.15 $\pm$ 2.9                                                                                    \\
\textbf{\method$_{Reg}$~+ Reg.}        & 44.92 $\pm$ 2.0                                                              & \textbf{28.85 $\pm$ 2.1}                                                                       & \textbf{3.36 $\pm$ 1.2}                                              & \textbf{89.33 $\pm$ 1.2}                                                                                    \\
 \midrule
               & \multicolumn{4}{c}{\textit{GHC $\to$ Stormfront}}                                                                                                                                         \\ \midrule
\textbf{Expl. Reg.}     & 51.53   $\pm$ 1.8                 &  13.43 $\pm$ 1.5                                         & \textbf{3.80    $\pm$ 0.4}                                                                              & 83.73 $\pm$ 8.0                                              \\
\textbf{Van-Transfer + Reg.}     &  52.18  $\pm$ 1.3                 & \textbf{13.12 $\pm$ 1.1}                                         & 4.35    $\pm$ 0.3                                                                              & 80.54 $\pm$ 2.0                                              \\
\textbf{\method$_{Reg}$~+ Reg.}       & \textbf{53.58 $\pm$ 1.4}                                           & 16.07 $\pm$ 1.3                                                       & 4.53 $\pm$ 0.9                             & \textbf{82.59 $\pm$ 1.4}                                                    \\
 \bottomrule
\end{tabular}
}
\caption{\textbf{Applying both UBM and downstream bias-mitigation} (\method$_{Reg}$ + Reg.), compared to downstream bias mitigation only (Expl. Reg.) and downstream bias mitigation over Van-Transfer model (Van-Transfer + Reg.).}
\label{tab:ubm_with_downstream}

\end{table}

Table~\ref{tab:same_domain_full} show the results of same-domain transfer with a single bias factors. Table~\ref{tab:mtl_same_full} further show the results of addressing multiple bias factors in this setup.

On GHC, Stormfront, and BiasBios, \method~overall reduces bias compared to Vanilla and Vanilla-Transfer. We notice the NYT accuracy on Stormfront in Stf. A $\to$ Stf. B setup is an exception. However, we see the bias is not reduced on Stf. B even when we directly run explanation regularization in the target domain. We reason that the Half-Stormfront dataset is small and the average length of the sentences are quite different between Stormfront and NYT, so that a model trained on Stormfront hardly generalizes to NYT.

We find intriguing results on FDCL; From FDCL A $\to$ FDCL B in Table~\ref{tab:same_domain_full}, we find bias is not reduced with~\method. However, as shown in Table~\ref{tab:mtl_same_full}, when the upstream model is trained jointly with other datasets to reduce multiple bias factors (Stf A + FDCL A, GHC A + FDCL A, GHC A + FDCL A + BiasBios A), the bias is clearly reduced.

\section{Applying UBM with Downstream Bias Mitigation}

In Table~\ref{tab:ubm_with_downstream}, we report the performance of performing both upstream and downstream bias mitigation, compared with downstream bias mitigation only, and downstream bias mitigation over a vanilla-transferred model. We see UBM further reduced bias in the Stormfront $\to$ GHC setup, while fail to improve in GHC $\to$ Stormfront. Compared to our previous results in Tables~\ref{tab:cross_full} and~\ref{tab:mtl_full}, we see a clearer directionality of transfer of bias mitigation effects when downstream bias mitigation is also applied.
\end{document}